# A Study on the Appropriate Size of the Mongolian General Corpus

Choi Sun Soo[1] and Ganbat Tsend[2]

[1]University of the Humanities, Ulaanbaatar, Mongolia
[2]Otgontenger University, Ulaanbaatar, Mongolia

*ABSTRACT*

*This study aims to determine the appropriate size of the Mongolian general corpus. This study used the Heaps' function and Type-Token Ratio (TTR) to determine the appropriate size of the Mongolian general corpus. This study's sample corpus of 906,064 tokens comprised texts from 10 domains of newspaper politics, economy, society, culture, sports, world articles and laws, middle and high school literature textbooks, interview articles, and podcast transcripts. First, we estimated the Heaps' function with this sample corpus. Next, we observed changes in the number of types and TTR values while increasing the number of tokens by one million using the estimated Heaps' function. As a result of observation, we found that the TTR value hardly changed when the number of tokens exceeded 39~42 million. Thus, we conclude that an appropriate size for a Mongolian general corpus is 39-42 million tokens.*

## 1. INTRODUCTION

The importance of the well balanced and representative general corpus of a language cannot be overstated. The general corpus plays an inseparable and important role not only in language research but also in artificial intelligence, natural language processing, and machine translation. Various general corpora have been composed in English and Korean, etc., but the Mongolian general corpus has not yet been composed.

In composing a general corpus, the size of the general corpus, along with its balance and representativeness, is one of the most important things to consider. With the development of computer, the size of the general corpus is growing day by day. However, is the larger the size of the general corpus, the better? Sinclair [1] stated, "The default value of Quantity is large."

However, the larger the size of the general corpus, the larger the amount of data to be processed, which reduces the effectiveness of analysis and utilization. And, since it takes a lot of time and money to compose a general corpus, it cannot be composed infinitely large. So we have no choice but to compromise on a reasonable general corpus' size.

Before composing a Mongolian general corpus, if there is an information about the appropriate size of the Mongolian general corpus, it will be of great help in composing the Mongolian general corpus. Then, what size of the Mongolian general corpus is appropriate for analysis and utilization? Finding the answer to this question is the purpose of this study.

International Journal on Natural Language Computing (IJNLC) Vol.12, No.3, June 2023

## 2. RELATED WORKS

In the meantime, in various countries and languages, general corpus of various sizes have been composed. And many studies have been conducted on the appropriate size of the general corpus. First, let's take a look at the size of some corpus composed in Mongolian, English, and Korean.

Table 1. Sizes* of various corpus.

| Language | Corpus | Year | Size | Reference |
|---|---|---|---|---|
| Mongolian | Language Resources for Mongolian | 2011 | 5,090,270 | [2] |
| English | Brown Corpus | 1964 | 1,014,312 | [3] |
| English | BNC (British National Corpus) | 1995 | 98,363,783 | [4] |
| English | COCA (Corpus of Contemporary American English) | 1990-present | 1,000,000,000 | [5] |
| Korean | Sejong Balanced Corpus | 2011 | 10,000,000 | [6] |

*Here, Size means the number of tokens of type I, but, only in BNC, means that of tokens of word-units (Type I + Type III, see section 3).

As shown in Table 1, the size of the general corpus varies greatly, from 1 million to 1 billion. This means that there is no set size for how large a general corpus should be. In fact, every general corpus is a sample of a population of a language, and no sample can be perfect.
Biber [7] proposed the following recursive method of composing a general corpus as shown in Figure 1.

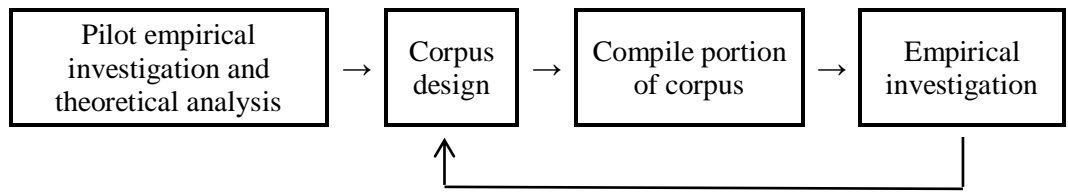

Figure 1. Biber's recursive method of corpus composition.

The key point of the method proposed by Biber is that an improved corpus should be composed recursively through various statistical analyses after creating one pilot corpus, rather than creating a perfect corpus from the beginning. This is also true for determining the appropriate size of the corpus. Therefore, the result of the present study is not perfect one and should be regarded as one research result that can serve as an example for determining the appropriate size of the Mongolian general corpus in the future.

Kyung Sook Yang et al. [8] suggested that the appropriate size of a Korean general corpus is about 10 million tokens using the cubic model, power model, and ARIMA (autoregressive integrated moving average) model. Sejong balanced corpus was composed with a size of 10 million tokens, which is in line with the results of this study.

Yukie Sano et al. [9] showed that the number of types can be predicted according to the size of the corpus using the estimated Zipf's and Heaps' function. We will also use the Heaps' function to estimate the number of potential types as the number of tokens increases.

## 3. DEFINITION OF TERMS

The definitions of key terms used in this study are as follows.

**General Corpus:** McEnery et al. [10] defined a corpus as follows.

A corpus is a collection of machine readable, authentic texts (including transcripts of spoken data) which is sampled to be representative of a particular language or language variety.

According to McEnery et al.'s definition, we defined a general corpus as a collection of machine readable, authentic texts (including transcripts of spoken data) which is sampled to be representative of a particular language. Therefore, General corpus is not specifically restricted to any particular subject field, register, domain or genre.

After Peirce [11] first introduced the concepts of type, token, and tone to science, type and token became widely used in natural language processing. In this study, the type is divided into type I, II, and III and defined as follows.

**Type:** Type is a form that exists in common in certain tokens, and refers to different forms among the tokens.
**Type I:** Type I refers to the different forms of a sequence of letters and numbers separated by spaces on both sides in a raw corpus. It can also be called word form.
**Type II:** Type II refers to different forms among the forms in which Type I is converted into the root form. It usually refers to the form of a headword in a dictionary. It can also be called lemma.
**Type III:** Type III refers to a unit that consists of two or more tokens and forms one meaning. It can also be called a multi-word unit.
For better understanding, we will show an example of preprocessing the corpus with type I, II, and III.

Table 2. Example of preprocessing a corpus with type I, II and III.

| Raw corpus |
|---|
| Гучин гуравдугаар зүйл.<br>1/Улсын Их Хурлын баталсан хууль, бусад шийдвэрт бүхэлд нь буюу зарим хэсэгт нь хориг тавих.<br><br>Article Thirty-three.<br>1/to exercise right to veto, either partially or wholly, against laws and other decisions adopted by the National Assembly of Mongolia. |
| Example of preprocessing with type I (word form) |
| Гучин гуравдугаар зүйл.<br>1 **Улсын Их Хурлын баталсан** хууль бусад **шийдвэрт бүхэлд** нь буюу зарим **хэсэгт** нь хориг тавих. |
| Example of preprocessing with type II (lemma) |
| Гучин гуравдугаар зүйл.<br>1 **Улс Их Хурал батлах** хууль бусад **шийдвэр бүхэл** нь буюу зарим **хэсэг** нь хориг тавих. |
| Example of preprocessing with type III (multi-word unit) |
| **Гучингуравдугаар** зүйл.<br>1 **Улсыниххурал** батлах хууль бусад шийдвэр бүхэл нь буюу зарим хэсэг нь хориг тавих. |

**Token:** Token is a sequence of letters or numbers separated by spaces on both sides in the corpus. A token contains a type.

19

**Size of corpus:** The number of all tokens in corpus.

**Appropriate size of corpus:** In this study, Appropriate size of corpus means the number of tokens at the point where the number of types (in this study type I) hardly increases even if the number of tokens increases. An increase of the TTR value is taken as a criterion for this.

## 4. METHODS

To estimate the appropriate size of the Mongolian general corpus, we used the following methods.

a) To compose a sample corpus of various domains (registers) of the Mongolian language.
b) To make the size of all corpus for each domain to be the same by random sampling.
c) To estimate Heaps' function.
d) To calculate Type-Token Ratio (TTR).
e) To find the appropriate size of the corpus.

Now, we will explain each method in detail.

### 4.1. Composition of a Sample Corpus

For this study, we composed a sample corpus by 10 domains as Table 3.

Table 3. Sample corpus.

| Corpus by domain | How to compose | Size (Tokens) |
|---|---|---|
| C1 (Newspaper-Culture) | Articles downloaded from Mongolian daily newspapers, Өнөөдөр, Өдрийн сонин, Зууны мэдээ, are compiled. 2019. 1.1. ~ 2019. 12. 31. | Raw: 135,384 Sample: 90,605 |
| C2 (Newspaper-Sports) | Articles downloaded from Mongolian daily newspapers, Өнөөдөр, Өдрийн сонин, Зууны мэдээ, are compiled. 2019. 1.1. ~ 2019. 12. 31. | Raw: 177,793 Sample: 90,605 |
| C3 (Newspaper-World) | Articles downloaded from Mongolian daily newspapers, Өнөөдөр, Өдрийн сонин, Зууны мэдээ, are compiled. 2019. 1.1. ~ 2019. 12. 31. | Raw: 155,097 Sample: 90,605 |
| C4 (Law) | All Mongolian laws as of 2020. 01. 03 were downloaded from the https://www.legalinfo.mn. | Raw: 1,634,714 Sample: 90,607 |
| C5 (Newspaper-Politics) | From the Mongolian daily newspaper 'Өдрийн сонин' website, from January 1, 2019 to December 31, 2019, a day's worth of articles randomly selected by week, a total of 52 day's articles were downloaded. | Raw: 94,830 Sample: 90,605 |
| C6 (Newspaper-Economy) | All articles from the Mongolian daily newspaper 'Өдрийн сонин' website from January 1, 2019 to December 31, 2019, among the articles in 'Өнөөдөр', a day's worth of articles randomly selected by week, a total of 52 days downloaded. The period for extracting articles from 'Өнөөдөр' is also 2019. 1.1. ~ 2019. 12. 31. | Raw: 109,312 Sample: 90,609 |
| C7 (Newspaper-Society) | From the Mongolian daily newspaper 'Өдрийн сонин' website, from January 1, 2019 to December 31, 2019, a day's worth of articles randomly selected by week, a total of 52 day's articles were downloaded. | Raw: 214,798 Sample: 90,610 |

| C8 (Textbook-Literature) | In 2019, all the works in Mongolian 6th-12th grade literature textbooks were directly entered into txt files. | Raw: 90,605 Sample: 90,605 |
|---|---|---|
| C9 (Interview) | All interview articles conducted with 143 Mongolians from 2010 to 2020 of gogo café on the 'Gogo.mn' site were downloaded. | Raw: 410,525 Sample: 90,608 |
| C10 (Podcast transcription) | A total of 29 hours and 59 minutes of 36 podcast broadcast recording files from 'https://soundcloud.com/caak-podcast/sets' were downloaded, and transcribed. | Raw: 255,221 Sample: 90,605 |

The corpus with the smallest size was the corpus of literature textbooks. Therefore, the sample size of corpus of other domains was set to 90,605, which is the same as the corpus of literature textbooks, by random sampling method. The number of tokens in the above corpus was calculated using the AntConc program [12]. Arabic numerals were excluded when calculating the number of tokens.

### 4.2. Heaps' Function (a.k.a. Heaps' Law)

Heaps [13] proposed the following exponential relationship between the number of types and the number of tokens in the corpus. The Heaps' function is:

$$V = k \cdot N^\beta \qquad (1)$$

where:
- $V$: Total number of types in the corpus (type I);
- $N$: Total number of tokens in the corpus;
- $k$: Heaps' coefficient;
- $\beta$: Heaps' exponent.

The Heaps' function expresses the exponential relationship between the number of tokens and the number of types in a corpus. Using the Heaps' function, we can predict the number of types according to the size of the corpus, that is, the number of tokens. There is one thing to note here. That is, the number of types and the number of tokens depends on the definition of the type. Therefore, In corpus-based research, the definition of type is very important. Depending on the definition of type, the results of corpus-based studies vary. Therefore, in corpus-based research, defining what a type is is one of the most important things. In this study, type and token are defined as section 3. In this study, type means type I, that is, word form. Therefore, $V$ in the Heaps' function means the total number of type I, that is, different word forms, neither type II (lemma) nor type III (multi-word unit).

### 4.3. Estimating the Heaps' Function

The process of estimating the Heaps' function is as follows.

a) Calculate the number of types and tokens for each sample corpus in Table 3.
b) Arrange the corpus in order of the corpus with the largest number of types to the corpus with the smallest number of types.
c) Calculate the number of cumulative tokens and types in the order listed in b).
d) Calculate the logarithm of the cumulative number of types and the number of tokens.
e) Obtain a linear regression equation using the logarithm of the type and number of tokens.

21

f) Convert the linear regression equation to a Heaps' function. Let's call it Heaps' function 1.
g) Randomly shuffle the order of the corpus to find a new Heaps' function and let's call call it Heaps' function 2.

We will estimate the value of *V* while changing the value of *N* with the estimated Heaps' function 1 and 2. Then, TTR will be estimated using the estimated value of *V*.

### 4.4. Calculating Type-Token Ratio (TTR)

The TTR is:

$$TTR = \frac{Total\ number\ of\ types}{Total\ number\ of\ tokens} = \frac{V}{N} \tag{2}$$

If *V* is almost unchanged no matter how large *N* is, there will be little change in TTR value. In other words, the point where the change in the TTR value becomes small can be said to be the appropriate size of the corpus. No matter how large *N* is, if *V* does not change, there is no need to make the corpus larger. Therefore, when estimating the appropriate size of the corpus, it is good to consider the amount of change in the TTR value as well.

### 4.5. Finding the appropriate size of the corpus

The process of determining the appropriate size of the corpus is as follows.
First, Estimate *V* with the estimated Heaps' function 1 and 2. And obtain estimated TTR.
Second, Find the point, *N*, at which the increase in TTR is smaller than 0.0001. Why 0.0001? There is no scientific, objective basis for this. However, this value was set as a reference value on the basis that the amount of change in TTR should be close to zero.

## 5. EMPIRICAL ANALYSIS AND RESULTS

### 5.1. Data Sets for Heaps' Function Estimation

In this study, the Heaps' function was estimated by the two methods mentioned in section 4.3. The two data sets used for estimating the Heaps' function are as Table 4.

Table 4. Two data sets for the Heaps' function.

|  | Sub Corpus | Token | Type I |
|---|---|---|---|
| Data set 1 | C8 | 90,605 | 19,765 |
|  | C8+C1 | 181,210 | 31,546 |
|  | C8+C1+C3 | 271,815 | 40,394 |
|  | C8+C1+C3+C9 | 362,423 | 45,933 |
|  | C8+C1+C3+C9+C2 | 453,028 | 51,364 |
|  | C8+C1+C3+C9+C2+C7 | 543,638 | 55,596 |
|  | C8+C1+C3+C9+C2+C7+C6 | 634,247 | 58,809 |
|  | C8+C1+C3+C9+C2+C7+C6+C10 | 724,852 | 63,380 |
|  | C8+C1+C3+C9+C2+C7+C6+C10+C5 | 815,457 | 65,113 |
|  | C8+C1+C3+C9+C2+C7+C6+C10+C5+C4 | 906,064 | 66,101 |
| Data set 2 | C1 | 90,605 | 17,903 |
|  | C1+C2 | 181,210 | 25,899 |

International Journal on Natural Language Computing (IJNLC) Vol.12, No.3, June 2023

| | | | |
|---|---|---|---|
| | C1+C2+C3 | 271,815 | 34,496 |
| | C1+C2+C3+C4 | 362,422 | 36,582 |
| | C1+C2+C3+C4+C5 | 453,027 | 39,453 |
| | C1+C2+C3+C4+C5+C6 | 543,636 | 43,047 |
| | C1+C2+C3+C4+C5+C6+C7 | 634,246 | 46,830 |
| | C1+C2+C3+C4+C5+C6+C7+C8 | 724,851 | 57,601 |
| | C1+C2+C3+C4+C5+C6+C7+C8+C9 | 815,459 | 61,635 |
| | C1+C2+C3+C4+C5+C6+C7+C8+C9+C10 | 906,064 | 66,101 |

Data set 1 is the data obtained from the number of accumulated tokens and types in the order of the corpus with the largest number of types to the corpus with the smallest number of types. Data set 2 is the cumulative number of tokens and the number of types obtained after classifying the order of the corpus into written and spoken language and then randomly arranging the rest. The reason why the data set is divided into two is to find out how the Heaps' function varies depending on the data set. The graphs of the data sets in Table 4 are as follows.

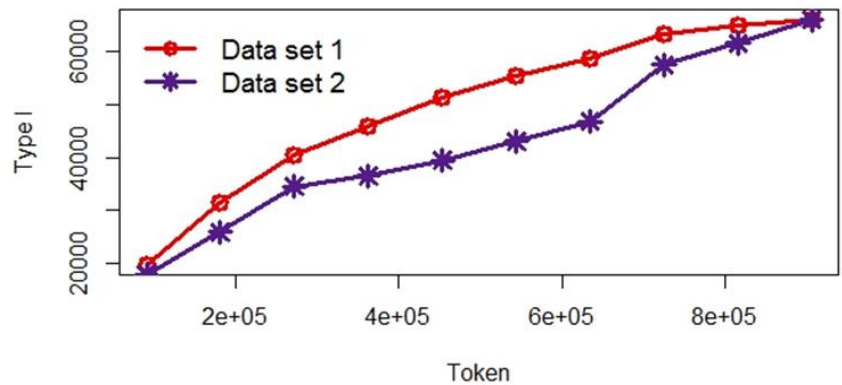

Figure 2. Two Graphs of two data sets.

Now, we will estimate two Heaps' functions using the two data sets in Table 4.

## 5.2. Heaps' Function Estimation

The two Heaps' functions estimated using the R with the data sets in Table 4 are as follows.

Table 5. Two Heaps' functions.

| Heaps' function 1 | Heaps' function 2 |
|---|---|
| $V = 56.31101 \cdot N^{0.52054}$ | $V = 35.40312 \cdot N^{0.5442}$ |

Table 5 confirms that the two Heaps' functions have almost the same $\beta$ value, but a large difference in $k$ value. Therefore, when estimating the Heaps' function, how the data is structured is very important. We will use these two functions to find the appropriate size of the corpus, respectively.

## 5.3. Estimation of *V* and TTR

Now, using the two Heaps' functions estimated in 5.2., the value of *V* and TTR according to the size of *N* by million are estimated as Appendices. The estimated values using Heaps' function 1 are in Appendix 1, and the estimated values using Heaps function 2 are in Appendix 2.

Although Heaps' function 1 and 2 are different, it can be seen that the TTR estimates of both functions show little change as *N* becomes larger than 40 million. The graph of the estimated values of *V* by the two Heaps' functions is shown in Figure 3.

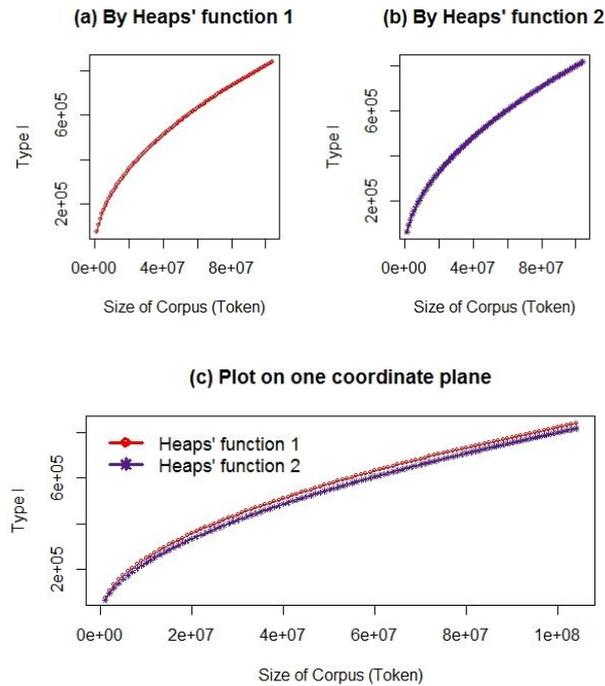

Figure 3.  Two Graphs of Estimated *V* (Type I) using the Heaps' function 1 and 2.

Figure 3 shows that as *N* increases, *V* also increases, but the amount of increase gradually decreases. To confirm this more accurately, TTR was obtained. The result is shown in Figure 4.

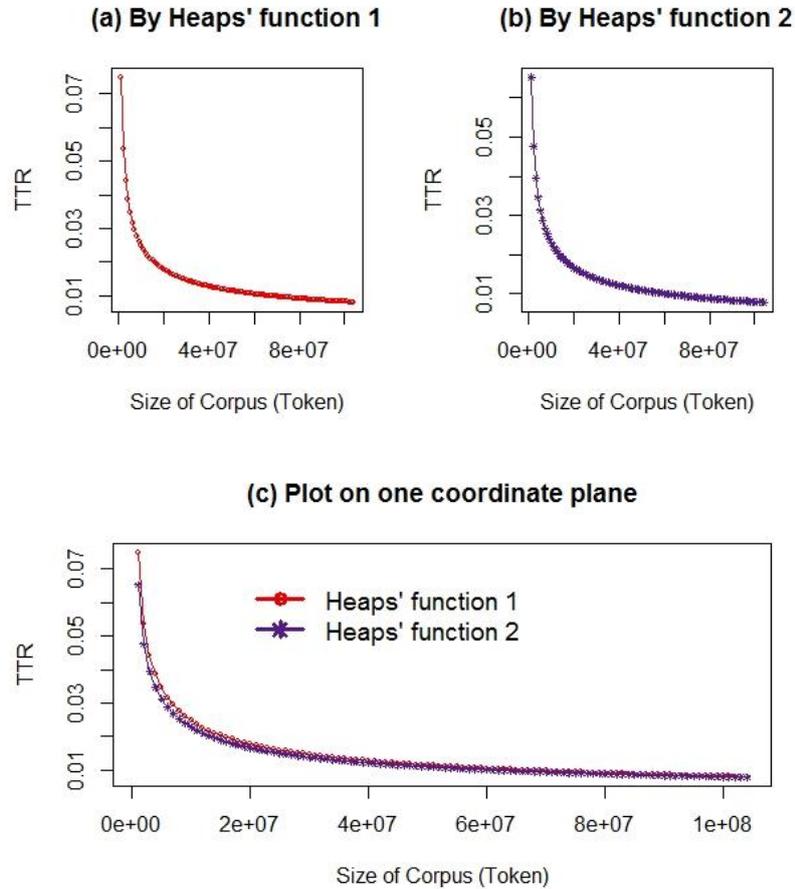

Figure 4. Two Graphs of Estimated TTR using the Heaps' function 1 and 2.

As shown in Figure 4, as *N* increases, TTR decreases rapidly at first, but after *N* becomes 40 million or more, it is almost unchanged. This means that new types rarely appear when *N* is over 40 million. Therefore, if the size of the Mongolian general corpus is about 40 million, it can be said that it contains sufficient types.

For a more accurate analysis, let's find the amount of change in TTR. The amount of change in TTR means the following.

$$\Delta TTR_{\text{n million}} = TTR_{\text{n million}} - TTR_{\text{n-1 million}} \qquad (3)$$

For example, subtracting the TTR when the number of tokens is 4 million from the TTR when the number of tokens is 5 million is the amount of change in TTR when the number of tokens is 5 milliom. The graph of the amount of change in TTR is shown in Figure 5.

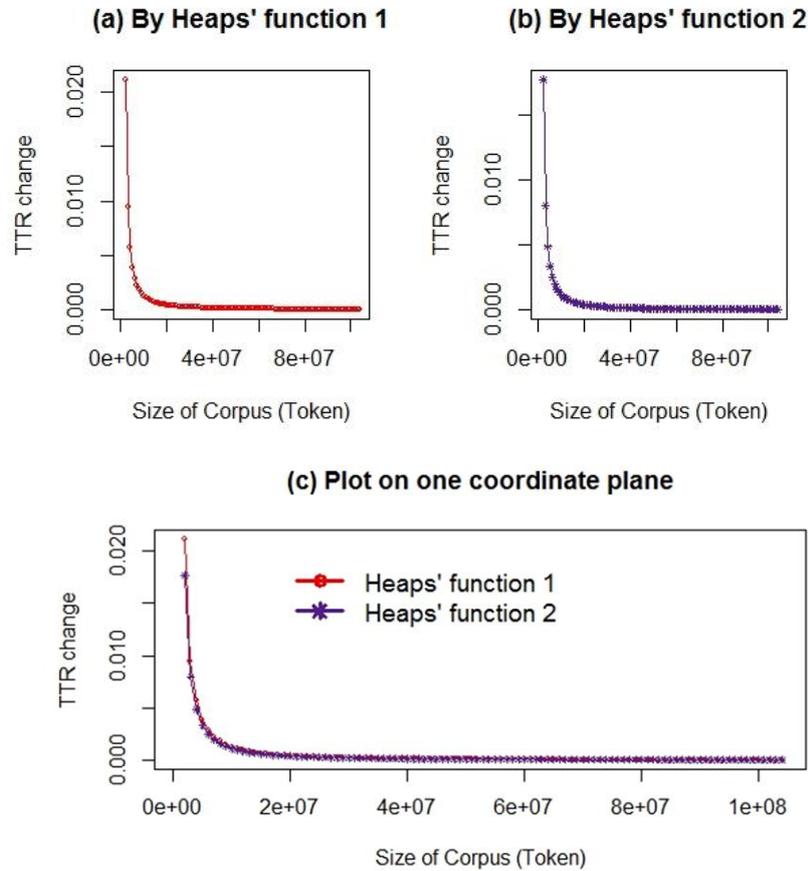

Figure 5. Two Graphs of Estimated TTR change using the Heaps' function 1 and 2.

In order to find the point where the amount of change in TTR becomes smaller than 0.0001, a part of the amount of change in TTR is shown in a Table 6 as follows.

Table 6. Estimated *V* and TTR change using the Heaps' function 1 and 2.

| Token (N) | By Heaps' function 1 | | | By Heaps' function 2 | | |
|---|---|---|---|---|---|---|
| | Type (V) | TTR | TTR change | Type (V) | TTR | TTR change |
| · | · | · | · | · | · | · |
| · | · | · | · | · | · | · |
| 37 million | 489942 | 0.0132 | 0.0002 | 465218 | 0.0126 | 0.0002 |
| 38 million | 496791 | 0.0131 | 0.0002 | 472019 | 0.0124 | 0.0002 |
| **39 million** | 503554 | 0.0129 | 0.0002 | 478739 | 0.0123 | **0.0001** |
| 40 million | 510234 | 0.0128 | 0.0002 | 485381 | 0.0121 | 0.0001 |
| 41 million | 516835 | 0.0126 | 0.0002 | 491947 | 0.012 | 0.0001 |
| **42 million** | 523359 | 0.0125 | **0.0001** | 498441 | 0.0119 | 0.0001 |
| 43 million | 529809 | 0.0123 | 0.0001 | 504865 | 0.0117 | 0.0001 |
| · | · | · | · | · | · | · |
| · | · | · | · | · | · | · |

In Table 6, the point at which the change in TTR becomes smaller than 0.0001 is when the number of tokens is 39 million for the heaps function 2 and 42 million for the heaps function 1.

Therefore, it can be said that the appropriate size of a Mongolian general corpus is between 39 and 42 million.

## 6. CONCLUSION

In order to predict the appropriate size of the Mongolian general corpus, this study composed a sample corpus of 906,064 tokens in 10 Mongolian language domains. Two Heaps' functions were estimated using the composed sample corpus. The first Heaps' function was estimated after arranging the number of types for each 10 domains in order from the largest to the smallest. The first Heaps' function's *k* was 56.31101 and *β* was 0.52054. The second Heaps' function was estimated after discriminating only the written and spoken domain and randomly arranging the rest. The second Heaps' function's *k* was 35.40312 and *β* was 0.5442.

The values of *V* and TTR were estimated while increasing *N* by a million units using the estimated two Heaps' functions. As a result, in the case of Heaps' function 1, when *N* is 42 million, and in the case of Heaps' function 2, when *N* is 39 million, the change in TTR starts to become less than 0.0001. Therefore, this study concludes that the appropriate size of a Mongolian general corpus is between 39 and 42 million.

The above results are valid results only in the present study based on the sample corpus used in this study. Therefore, the results of studies using other sample corpus may come out differently. However, it is meaningful that this study presented an example of the appropriate size of the Mongolian general corpus. In the future, we expect to draw more elaborate conclusions by using more diverse functions based on various sample corpus.

Sinclair, J. (1996). Preliminary Recommendations on Corpus Typology. EAGLES Document EAG-TCWG-CTYP/P. Available at:
http://www.ilc.cnr.it/EAGLES96/corpustyp/corpustyp.html. [Accessed 13/10/2011]

### REFERENCES

[1] John, Sinclair. (1996). Preliminary recommendations on corpus typology, EAGLES document, EAG-TCWG-CTYP/P, version of May, p. 6.
[2] Purev J., Altangerel Ch. (2011). Language resources for Mongolian. *Conference on Humane Language Technology for Development*, Alexandria, Egypt.
[3] Francis W. Nelson., and Henry Kučera. (1964, 1971, 1979). *MANUAL OF INFORMATION to accompany: A Standard Corpus of Present-Day Edited American English, for use with Digital Computers*. Rhode island, USA: Brown University.
[4] Lou, Burnard. (2007). *Reference Guide for the British National Corpus (XML Edition).* Published for the British National Corpus Consortium by the Research Technologies Service at Oxford University Computing Services. Available at: http://www.natcorp.ox.ac.uk/docs/URG/
[5] Mark, Davies. (2010). The Corpus of Contemporary American English as the first reliable monitor corpus of English. *Literary and Linguistic Computing*, Vol. 25, No. 4. pp. 447-464. Available at: https://www.english-corpora.org/davies/articles/davies_44.pdf
[6] Hwang Yong Ju., Choi Jung Do. (2016). 21 세기 세종 말뭉치 제대로 살펴보기- 언어정보 나눔터 활용하기, 새국어 생활 (New Korean Life), Vol. 26, No. 2, 국립국어원 (National Institute of Korean Language), pp. 73-86. Available at: https://www.korean.go.kr/nkview/nklife/2016_2/26_0204.pdf
[7] Douglas, Biber. (1993). Repesentativeness in Corpus Design, *Literary and Linguistic computing*, vol. 8, no. 4, Oxford University Press, p. 256.
[8] Kyung Sook Yang., Byung Sun Park., and Jin Ho Lim. (2003). The Statistical Relationship between Types and Tokens, *Annual Conference on Human and Language Technology*, Seoul, Korea, pp. 81-85. (pISSN: 2005-3053)

[9] Yukie Sano., Hideki Takayasu., and Misako Takayasu. (2012). "Zipf's Law and Heaps' Law Can Predict the Size of Potential Words", *Progress of Theoretical Physics Supplement,* No. 194, pp. 202-209.
[10] Tony, McEnery., Richard, Xiao., Yukio, Tono. (2005). *Corpus-based language studies: An Advanced Resource Book*. Oxfordshire, UK: Routledge Taylor & Francis Group. p. 5.
[11] Charles S. Peirce. (1906). Prolegomena to an apology for pragmaticism. *The Monist, 16(4).* pp. 492–546. Available at: http://www.jstor.org/stable/10.2307/27899680
[12] Laurence, Anthony. (2019). *AntConc (Version 3.5.8) [Computer Software]*. Tokyo, Japan: Waseda University. (https://www.laurenceanthony.net/software)
[13] Harold, Stanley, Heaps. (1978). *Information Retrieval: Computational and Theoretical Aspects*. Academic Press, Orland. USA.

**AUTHORS**

**Choi Sun Soo**, is a doctoral student in Linguistics in the University of the Humanities. His areas of interest are: Corpus Linguistics, Corpus Composition, and Deep learning and Data Processing.

**Ganbat Tsend**, is a professor in Otgontenger University. His R&D interests are a natural language processing, machine learning, digital transition, system analysis and design. He is an author of several IT books as human and computer interaction, information retrieval, computer modeling and simulation.

**APPENDIX 1**

Estimated *V* and TTR using the Heaps' function 1.

| Token (N) | Type (V) | TTR | Token (N) | Type (V) | TTR | Token (N) | Type (V) | TTR |
|---|---|---|---|---|---|---|---|---|
| 1 million | 74788 | 0.0748 | 35 million | 475973 | 0.0136 | 69 million | 677685 | 0.0098 |
| 2 million | 107283 | 0.0536 | 36 million | 483004 | 0.0134 | 70 million | 682780 | 0.0098 |
| 3 million | 132493 | 0.0442 | 37 million | 489942 | 0.0132 | 71 million | 687840 | 0.0097 |
| 4 million | 153897 | 0.0385 | 38 million | 496791 | 0.0131 | 72 million | 692866 | 0.0096 |
| 5 million | 172852 | 0.0346 | 39 million | 503554 | 0.0129 | 73 million | 697859 | 0.0096 |
| 6 million | 190061 | 0.0317 | 40 million | 510234 | 0.0128 | 74 million | 702819 | 0.0095 |
| 7 million | 205940 | 0.0294 | 41 million | 516835 | 0.0126 | 75 million | 707746 | 0.0094 |
| 8 million | 220764 | 0.0276 | 42 million | 523359 | 0.0125 | 76 million | 712643 | 0.0094 |
| 9 million | 234722 | 0.0261 | 43 million | 529809 | 0.0123 | 77 million | 717509 | 0.0093 |
| 10 million | 247955 | 0.0248 | 44 million | 536187 | 0.0122 | 78 million | 722344 | 0.0093 |
| 11 million | 260567 | 0.0237 | 45 million | 542496 | 0.0121 | 79 million | 727150 | 0.0092 |
| 12 million | 272640 | 0.0227 | 46 million | 548738 | 0.0119 | 80 million | 731927 | 0.0091 |
| 13 million | 284240 | 0.0219 | 47 million | 554916 | 0.0118 | 81 million | 736675 | 0.0091 |
| 14 million | 295419 | 0.0211 | 48 million | 561031 | 0.0117 | 82 million | 741396 | 0.009 |
| 15 million | 306222 | 0.0204 | 49 million | 567085 | 0.0116 | 83 million | 746088 | 0.009 |
| 16 million | 316684 | 0.0198 | 50 million | 573080 | 0.0115 | 84 million | 750754 | 0.0089 |
| 17 million | 326837 | 0.0192 | 51 million | 579018 | 0.0114 | 85 million | 755393 | 0.0089 |
| 18 million | 336708 | 0.0187 | 52 million | 584900 | 0.0112 | 86 million | 760006 | 0.0088 |
| 19 million | 346319 | 0.0182 | 53 million | 590728 | 0.0111 | 87 million | 764594 | 0.0088 |
| 20 million | 355690 | 0.0178 | 54 million | 596504 | 0.011 | 88 million | 769156 | 0.0087 |
| 21 million | 364839 | 0.0174 | 55 million | 602229 | 0.0109 | 89 million | 773693 | 0.0087 |
| 22 million | 373782 | 0.017 | 56 million | 607904 | 0.0109 | 90 million | 778206 | 0.0086 |
| 23 million | 382531 | 0.0166 | 57 million | 613531 | 0.0108 | 91 million | 782695 | 0.0086 |
| 24 million | 391101 | 0.0163 | 58 million | 619110 | 0.0107 | 92 million | 787161 | 0.0086 |
| 25 million | 399500 | 0.016 | 59 million | 624644 | 0.0106 | 93 million | 791603 | 0.0085 |
| 26 million | 407740 | 0.0157 | 60 million | 630133 | 0.0105 | 94 million | 796022 | 0.0085 |
| 27 million | 415830 | 0.0154 | 61 million | 635578 | 0.0104 | 95 million | 800419 | 0.0084 |
| 28 million | 423777 | 0.0151 | 62 million | 640981 | 0.0103 | 96 million | 804794 | 0.0084 |
| 29 million | 431589 | 0.0149 | 63 million | 646342 | 0.0103 | 97 million | 809147 | 0.0083 |
| 30 million | 439272 | 0.0146 | 64 million | 651662 | 0.0102 | 98 million | 813479 | 0.0083 |
| 31 million | 446835 | 0.0144 | 65 million | 656942 | 0.0101 | 99 million | 817789 | 0.0083 |
| 32 million | 454280 | 0.0142 | 66 million | 662184 | 0.01 | 100 million | 822079 | 0.0082 |
| 33 million | 461616 | 0.014 | 67 million | 667388 | 0.01 | 101 million | 826348 | 0.0082 |
| 34 million | 468845 | 0.0138 | 68 million | 672555 | 0.0099 | 102 million | 830596 | 0.0081 |

**APPENDIX 2**

Estimated *V* and TTR using the Heaps' function 2.

| Token (N) | Type (V) | TTR | Token (N) | Type (V) | TTR | Token (N) | Type (V) | TTR |
|---|---|---|---|---|---|---|---|---|
| 1 million | 65199 | 0.0652 | 35 million | 451360 | 0.0129 | 69 million | 653045 | 0.0095 |
| 2 million | 95074 | 0.0475 | 36 million | 458333 | 0.0127 | 70 million | 658179 | 0.0094 |
| 3 million | 118547 | 0.0395 | 37 million | 465218 | 0.0126 | 71 million | 663279 | 0.0093 |
| 4 million | 138638 | 0.0347 | 38 million | 472019 | 0.0124 | 72 million | 668347 | 0.0093 |
| 5 million | 156538 | 0.0313 | 39 million | 478739 | 0.0123 | 73 million | 673383 | 0.0092 |
| 6 million | 172867 | 0.0288 | 40 million | 485381 | 0.0121 | 74 million | 678387 | 0.0092 |
| 7 million | 187994 | 0.0269 | 41 million | 491947 | 0.012 | 75 million | 683361 | 0.0091 |
| 8 million | 202164 | 0.0253 | 42 million | 498441 | 0.0119 | 76 million | 688304 | 0.0091 |
| 9 million | 215546 | 0.0239 | 43 million | 504865 | 0.0117 | 77 million | 693218 | 0.009 |
| 10 million | 228266 | 0.0228 | 44 million | 511221 | 0.0116 | 78 million | 698103 | 0.009 |
| 11 million | 240418 | 0.0219 | 45 million | 517511 | 0.0115 | 79 million | 702959 | 0.0089 |
| 12 million | 252076 | 0.021 | 46 million | 523738 | 0.0114 | 80 million | 707788 | 0.0088 |
| 13 million | 263299 | 0.0203 | 47 million | 529904 | 0.0113 | 81 million | 712589 | 0.0088 |
| 14 million | 274135 | 0.0196 | 48 million | 536010 | 0.0112 | 82 million | 717363 | 0.0087 |
| 15 million | 284623 | 0.019 | 49 million | 542059 | 0.0111 | 83 million | 722111 | 0.0087 |
| 16 million | 294797 | 0.0184 | 50 million | 548051 | 0.011 | 84 million | 726832 | 0.0087 |
| 17 million | 304686 | 0.0179 | 51 million | 553989 | 0.0109 | 85 million | 731529 | 0.0086 |
| 18 million | 314312 | 0.0175 | 52 million | 559874 | 0.0108 | 86 million | 736200 | 0.0086 |
| 19 million | 323697 | 0.017 | 53 million | 565708 | 0.0107 | 87 million | 740846 | 0.0085 |
| 20 million | 332860 | 0.0166 | 54 million | 571492 | 0.0106 | 88 million | 745468 | 0.0085 |
| 21 million | 341817 | 0.0163 | 55 million | 577227 | 0.0105 | 89 million | 750066 | 0.0084 |
| 22 million | 350581 | 0.0159 | 56 million | 582915 | 0.0104 | 90 million | 754641 | 0.0084 |
| 23 million | 359165 | 0.0156 | 57 million | 588557 | 0.0103 | 91 million | 759192 | 0.0083 |
| 24 million | 367581 | 0.0153 | 58 million | 594154 | 0.0102 | 92 million | 763721 | 0.0083 |
| 25 million | 375838 | 0.015 | 59 million | 599707 | 0.0102 | 93 million | 768228 | 0.0083 |
| 26 million | 383946 | 0.0148 | 60 million | 605217 | 0.0101 | 94 million | 772712 | 0.0082 |
| 27 million | 391913 | 0.0145 | 61 million | 610686 | 0.01 | 95 million | 777175 | 0.0082 |
| 28 million | 399747 | 0.0143 | 62 million | 616114 | 0.0099 | 96 million | 781616 | 0.0081 |
| 29 million | 407454 | 0.0141 | 63 million | 621502 | 0.0099 | 97 million | 786037 | 0.0081 |
| 30 million | 415041 | 0.0138 | 64 million | 626852 | 0.0098 | 98 million | 790436 | 0.0081 |
| 31 million | 422513 | 0.0136 | 65 million | 632163 | 0.0097 | 99 million | 794815 | 0.008 |
| 32 million | 429877 | 0.0134 | 66 million | 637437 | 0.0097 | 100 million | 799174 | 0.008 |
| 33 million | 437136 | 0.0132 | 67 million | 642675 | 0.0096 | 101 million | 803514 | 0.008 |
| 34 million | 444296 | 0.0131 | 68 million | 647877 | 0.0095 | 102 million | 807833 | 0.0079 |